\RequirePackage{tikz} \usetikzlibrary{shapes,arrows}
\documentclass[pdflatex,sn-mathphys]{sn-jnl}
\usepackage{subcaption}
\usepackage{arydshln}
\jyear{2023}%

\theoremstyle{thmstyleone}%
\theoremstyle{thmstyletwo}%

\theoremstyle{thmstylethree}%

\newcommand{\Vx}{\textbf{Var}_x}
\newcommand{\Ex}{\textbf{E}_x}
\newcommand{\Vz}{\textbf{Var}_z}
\newcommand{\V}{\textbf{Var}}
\newcommand{\Ez}{\textbf{E}_z}
\newcommand{\bx}{\mathbf{x}}
\newcommand{\bz}{\mathbf{z}}

\begin{document}

\title[A moment-matching metric for latent variable generative models]{A moment-matching metric for latent variable generative models}

\author[1]{\fnm{C\'edric} \sur{Beaulac}}
\email{beaulac.cedric@gmail.com}

\affil[1]{\orgdiv{Département de mathématiques}, \orgname{Université du Québec à Montréal}, \orgaddress{\city{Montréal}, \state{Québec}, \country{Canada}}}

\abstract{It is difficult to assess the quality of a fitted model in unsupervised learning problems. Latent variable models, such as variational autoencoders and Gaussian mixture models, are often trained with likelihood-based approaches. In the scope of Goodhart's law, when a metric becomes a target it ceases to be a good metric and therefore we should not use likelihood to assess the quality of the fit of these models. The solution we propose is a new metric for model comparison or regularization that relies on moments. The key idea is to study the difference between the data moments and the model moments using a matrix norm, such as the Frobenius norm. We first show how to use this new metric for model comparison and then for regularization. We show that our proposed metric is faster to compute and has a smaller variance than the commonly used procedure of drawing samples from the fitted distribution. We conclude this article with a demonstration of both applications and we discuss our findings and future work.}




\maketitle

\section*{Statements and Declarations}

The author has no conflict of interest to declare.

\pagebreak

\section{Introduction} \label{intr}

When fitting supervised models, statisticians and computer scientists alike have come up with a variety of metrics in order to evaluate the quality of their predictions, from the simple mean-squared error to more general loss functions. However, in the context of unsupervised learning there is no direct measure of success and it can be difficult to assess the validity of the fitted model \cite{Hastie09}.

\bigskip

Assume an unsupervised learning context where we have a data set $S= \{\mathbf{x}_1,...,\mathbf{x}_N\}$ and the abstract goal of capturing the distribution $p(\bx)$. One way to train such model is to assume a distribution and then maximize the likelihood of the data set with respect to the parameters of the distribution. If multiple models are trained, they are usually compared using the likelihood as well. Goodhart's law \cite{goodhart84,strathern97} states that \textit{when a measure becomes a target, it ceases to be a good measure} and thus we should not strictly rely on the likelihood to evaluate models that were trained with the likelihood.

\bigskip

In this article we propose a new way to assess the quality of the fit of a large family of unsupervised models with respect to our abstract goal of capturing the distribution $p(\bx)$. In other words, we propose a new way to measure if an estimated distribution $\hat{p}(\bx)$ resembles the observed distribution $p_S(\bx)$. More precisely, we offer a diagnostic technique for parametric latent variable models such as Variational AutoEncoders (VAEs) \cite{Kingma13,Kingma17} and Gaussian Mixture Models (GMMs). Our proposed metric evaluates the quality of the \textit{fitted model}; we compare the learned parameters of the unsupervised model with the observed data distribution the model is trying to capture. We do so by building distinct moment estimators and comparing them. The main purpose of such metric is to provide a new way to compare the fit of multiple models from different families by assessing how well these models captured the first two moments of the data. Though capturing the first and second moment of a data set is not a sufficient condition to claim the trained model has perfectly captured the data distribution, it certainly is a necessary condition under some assumption we discuss later. 

\bigskip

In statistics and machine learning, Goodhart's law is often compared with the concept of overfitting. One popular way to circumvent model overfitting has been regularization. Consequently, we offer a second perspective on our new metric; it can be used for regularization. Our metric favours simple models and thus it can be easily integrated in the optimization procedure as a regularizer.

\bigskip

The technique we propose is fast to compute, works for a wide range of models and is built upon a rigorous mathematical formulation. It provides a new way to compare multiple models or regularize them and behaves similarly to previously used heuristic techniques.

\bigskip

In the next section, we establish the family of latent variable models that this metric was designed for. We then discuss related work in section \ref{rewo} and we introduced the moment estimators used in section \ref{moes}.  In section \ref{mega}, we present our metric and its implementation and next we demonstrate how it performs for model evaluation on simple examples in section \ref{expe1}. We then introduce the framework for the application of our metric for regularization in section \ref{regu} and we demonstrate how it behaves as regularizer in section \ref{expe2}. Finally, we discuss the limitations of our approach in section \ref{limit} before some concluding remarks in section \ref{conc}.

\section{Latent Variable Generative Models}\label{lvgm}

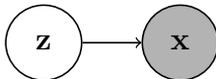
\begin{figure}[h!]
  \centering
  \begin{tikzpicture}[->, semithick,scale = 0.9]
  \tikzstyle{latent}=[fill=white,draw=black,text=black,style=circle,minimum size=1.0cm]
  \tikzstyle{observed}=[fill=black!30,draw=black,text=black,style=circle,minimum size=1.0cm]

  \node[latent]   (A) at (0,0)  {\large $\mathbf{z}$};

  \node[observed]         (C) at (2,0)  {\large $\mathbf{x}$};

  \path (A) edge            node {} (C);

\end{tikzpicture}
  \caption{Graphical representation of latent variables models with joint distribution $p(\mathbf{x},\mathbf{z}) = p(\mathbf{z})p(\mathbf{x}\|\bz)$.}
  \label{genny}
\end{figure}

Let us first define what we refer to as latent variable generative models (LVGMs). Assume we have a data set $S= \{\mathbf{x}_1,...,\mathbf{x}_N\}$ consisting of $N$ observations of a $D$-dimensional variable $\mathbf{x}$ where $\bx$ $\in$ $\mathcal{X}$ which is $D$-dimensional. We want to estimate the distribution of the random variable $\mathbf{x}$ but it is \textit{too complicated} to be captured by a simple distribution. Latent variable models suppose there exist an unobserved latent variable, say $\mathbf{z}$, that has a direct influence on the distribution of $\mathbf{x}$
\begin{align}
p(\bx) = \int_{z} p(\bx\| \bz)p(\bz) dz,
\label{LVMeq}
\end{align}
where we assume $\bz$ $\in$ $\mathcal{Z}$ which is $M$-dimensional. The model proposed by equation (\ref{LVMeq}) is quite general but allows relatively complex marginal distributions over observed variables $\mathbf{x}$ to be expressed in terms of more tractable conditional distributions $p(\bx\|\bz)$ \cite{Bishop07}. Similarly, it leads to a tractable joint distribution as well
\begin{align}
p(\bx,\bz) = p(\bz)p(\bx \| \bz),
\label{jointLVM}
\end{align}
and this is quite often represented using a simple graph as seen in Figure \ref{genny}.

These models are \textit{generative models} because learning $p(\bx,\bz)= p(\bz)p(\bx\|\bz)$ allows us to generate new samples of $\bx$ using ancestral sampling. What makes this model probabilistic is that the mapping from $\bz$ to $\bx$ is not a deterministic function $f: \mathcal{Z} \rightarrow \mathcal{X}$ but instead a \textit{probabilistic mapping} from $\mathcal{Z}$ to $\Theta$, where $\Theta$ is the parameter space of $p_\theta(\bx\|\bz)$; $\theta \in \Theta$. We call $p_\theta(\bx\|\bz)$ the emission distribution or observation distribution interchangeably.
 
\bigskip

When training or fitting such models, we train the function $f:\mathcal{Z} \rightarrow \Theta$ to maximize the likelihood of the data set $S$ under the model of equation (\ref{LVMeq}). This mapping $f$ explains the effect of $\bz$ on $\bx$ and is at the centre of latent variable models. Therefore learning this function $f$ is the main challenge of training latent variable models and the Expectation-Maximization algorithm (EM) or variational inference are common solutions to this problem. In most cases, $p(\bz)$ is assumed to be known and fixed but in some cases the parameters of $p(\bz)$ are estimated as well.

\bigskip

Usually $p_\theta(\bx\|\bz)$ is a simple parametric distribution and the latent variable increases the complexity of $p_\theta(\bx)$. Additionally, the function $f$ can take many forms, from simple linear combination to neural network functions. We use $f(\bz)$ interchangeably with $f$ or the \textit{distribution parameters} it outputs directly. 

\bigskip

Let us begin with a simple example. Assume the emission distribution is Poisson: $p_\theta(\bx\|\bz) = \text{Poisson}(\lambda)$ then $f: \mathcal{Z} \rightarrow \mathbb{R}^+$ because $\theta = \lambda \in \mathbb{R}^+$ and we use $f(\bz)$ and $\lambda(\bz)$ interchangeably. If there exist a simple mapping from the parameters of the distribution to its expectation and its variance, we also use them interchangeably. For the Poisson example, $\Ex[\bx\|\bz] = \lambda(\bz)$ and $ \Vx[\bx\|\bz] = \lambda(\bz)$. 

\bigskip

One important detail to bring up is that the moments are only meaningful for a certain family of emission distributions. For the application of our metric, we will consider the family of emission distribution for which the moment generating function exists. 

\subsection{Probabilistic Principal Component Analysis}

The Probabilistic Principal Component Analysis (pPCA) \cite{tipping99,Bishop07} is a member of the LVGM family we just described where $p(\bz)$ is assumed to be a normal distribution $N(0,I)$. The emission distribution is also assumed to be Normal: $p(\bx\|\bz) = N(W\bz+b,I\sigma^2)$. In this formulation, we see that $f: \mathbb{R}_M \rightarrow \mathbb{R}_D$ is a linear function that maps the latent variable $\bz$ to $\Ex[\bx\|\bz]$: $\Ex[\bx\|\bz] = \mu(\bz) = W\bz+b$ . Outside of estimating $W$ and $b$ as part of $f$, the model also estimates the parameter $\sigma^2$ though it is not a function of $\bz$. However it is a function of $D$ and $M$, the dimension of $\mathcal{X}$ and $\mathcal{Z}$. 

\bigskip

The parameters of pPCA can be obtained analytically as the solution of a direct maximization of the likelihood or with the EM algorithm.   

\subsection{Variational Autoencoders}

The VAE is also a member of the LVGM family. It is assumed that $\bz$ is a continuous variable where $p(\bz)$ is assumed to be $N(0,I)$ in the introductory papers \cite{Kingma13,Kingma17}. $p(\bx\|\bz)$ can be any parametric distribution where $f(\bz)$ outputs the parameters of this distribution. For instance, if $p(\bx\|\bz)$ is normal then $f(\bz)$ will output a mean and a variance parameter, $f: \mathbb{R}_M \rightarrow \mathbb{R}_D \times \mathbb{R}_{(D,D)}^+$.

\bigskip

One novelty of VAEs is that the function $f$ proposed is much more flexible than a linear combination; it is a neural network.  In turn, this makes the posterior distribution $p(\bz\|\bx)$ intractable and prevents the model from being fitted by the EM algorithm. The solution proposed is to assume a variational distribution $q(\bz\|\bx)$ and optimize the likelihood by maximizing the Evidence Lower BOund (ELBO) \cite{Bishop07,Kingma13}, a lower bound of the observed-data log-likelihood.

\subsection{Gaussian Mixture Models}

For a GMM with $K$-components we define $\bz$ as a $K$-class categorical variable with $\bz \in \{1,...,K\} = \mathcal{Z}$ and $p(\bz)$, a categorical distribution where  $\pi_j = p(\bz = j)$ and $\sum_{j=1}^K \pi_j =1$. Finally, setting $p(\bx\|\bz=j) = N(\mu_j,\Sigma_j)$ leads to a GMM:
\begin{align}
p(\bx) = \sum_{j=1}^K \pi_j p(\bx\|\bz=j). 
\label{LVM}
\end{align}

In this situation, $f$ maps the latent variable to a pair of \textit{distribution parameters}, $\mu$ and $\Sigma$, $f: \{1,...,K\} \rightarrow \mathbb{R}_{D}\times\mathbb{R}_{D \times D}^+$. In this particular case  $\Ex[\bx\|\bz] = f_1(\bz)$ where $f_1$ is the \textit{first} output of $f(\bz)$ or simply $\mu(\bz)$ and $\Vx[\bx\|\bz] = f_2(\bz) = \Sigma(\bz)$.

\bigskip

The GMM is a special case of LVGM where we also estimate the parameters $\{\pi_j : j \in 1,...K\}$ of $p(\bz)$, identifiable up to a permutation. A GMM is usually trained with the EM algorithm.

\section{Related Works} \label{rewo}

In this article we propose a new metric to evaluate the goodness-of-fit for the family of latent variable models defined in section \ref{lvgm} and in this section we discuss some common alternatives. When proposing new LVGMs researchers rely either on the likelihood of the data under the fitted model or a heuristic analysis of generated data points \cite{Kingma13,li15,higgins17,zhao19,vahdat20}. To evaluate the performance of the models, both those techniques have their fair share of problems which we discuss in this section. The metric we propose is an alternative to those techniques. 

\bigskip

A problem with evaluating models with the likelihood is that a high likelihood does not necessarily mean that the proposed model captured the distribution of the observed data. For instance, Bishop et al. \cite{Bishop07} demonstrate that for GMMs it is possible to have the likelihood be infinite ($\infty$) by setting the mean of one component to be exactly one of the observed points, say $\bx_i$, and then pushing the variance of that component to $0$, thus the likelihood of $\bx_i$ under that particular component will be infinite. Similarly, Zhao et al. \cite{zhao19} built a toy example where the ELBO (a lower bound of the log-likelihood) would converge to infinity.

\bigskip

Another commonly employed strategy is to generate new observations and try to determine if they look like real data with a simple visual inspection, this is a technique used by many authors \cite{Kingma13,li15,higgins17,Kingma17,zhao19,vahdat20}. A weakness of visual inspection is that it is subjective, it incentivizes cherry-picking of results and it is not a rigorous means of comparing models. 

\bigskip

Outside of papers that introduced new LVGMs, some techniques have been developed to compared samples from different distribution and to assess goodness-of-fit using kernels. The Maximum Mean Discrepancy (MMD) \cite{gretton12} is a measure of similarities between two samples which the authors used to construct statistical tests to determine if the two samples are drawn from different distributions. Unfortunately, this requires to sample from the fitted distribution, which we later demonstrate to be a slow procedure and the original implementation proposed by the authors is no longer available online. The Kernel Stein Discrepancy test \cite{Liu16} is a goodness-of-fit test that was later extended specifically for latent variable models \cite{kanagawa2019,jitkrittum2020}. Both of these Kernel Stein Discrepancy goodness-of-fit tests have strong theoretical guarantees. Unfortunately, these guarantees only hold under strict assumptions that can be complicated to verify. Additionally, the magnitude of this discrepancy metric is not only affected by the evaluated models by also the choice of reproducing kernel \cite{kanagawa2019}. On the opposite, our proposed metric works for any LVGMs that fit the definition of section \ref{lvgm} and is not affected by arbitrary choices. 

\bigskip

Comparing data moments with model moments have been proposed in the past but only for training purposes \cite{anandkumar12,anandkumar14b,chaganty14,li15,podosinnikova16}. Anandkumar \cite{anandkumar12,anandkumar14b} proposes efficient ways to code and optimize latent variable models by comparing the true data set with a sample generated from the model. Podosinnikova \cite{podosinnikova16} approaches this topic very thoroughly in their thesis where they also discuss the use of moment-generating function estimators. What we propose here is different; we propose a metric. Even though we discuss the possibility of using our metric for optimization in later sections we believe there already exists a rich literature that discusses new ways of optimizing latent variable models but few publications that address the lack of evaluation metrics.

\section{Moment Estimators} \label{moes}

In this section we define two different moment estimators for the first and the second moment. The goal is to build different estimators containing different information. To begin, we define moment estimators of the data set, we call those Data Estimators (DE). Then we define another set of moment estimators that represent the distribution captured by the LVGM which we call Forward Model Estimators (FME). 

\subsection{Second Moment}

We build two different estimators of the same quantity, the second moment. One uses observed data while the other uses the proposed generative model. What makes the proposed FME different is that we do not sample new data points from the LVGM but instead rely on a simple probability identity to build the FME.

\bigskip

To do so, let us introduce the well-know Law of Total Variance
\begin{align}
\Vx(\bx) = \Ez[\Vx(\bx\|\bz)] + \Vz[\Ex(\bx\|\bz)],
\end{align}
and notice that
\begin{align}
\Vz[\Ex(\bx\|\bz)] &= \Ez[\Ex(\bx\|\bz)^2] - (\Ez[\Ex(\bx\|\bz)])^2 \nonumber \\
&= \Ez[\Ex(\bx\|\bz)^2] - (\Ex[\bx])^2.
\end{align}

We combine and reorganize both equations
\begin{align}
\Vx(\bx)+ (\Ex[\bx])^2  = \Ez[\Vx(\bx\|\bz)] + \Ez[\Ex(\bx\|\bz)^2].
\label{rhslhs}
\end{align}
We have reorganized both terms in this particular was so that the left-hand side of equation (\ref{rhslhs}) is independent of the latent variables and can be estimated from $S$ independently from the choice of model while the right-hand side contains information about both the expectation and the variance of the generative model. Additionally, notice the left-hand side is actually $\Ex[\bx^2]$, the second moment of $\bx$ and thus our work here consists of comparing two different estimators of $\Ex[\bx^2]$ which we introduce next. The left-hand side of equation (\ref{rhslhs}) can be estimated using the observed data  
\begin{align}
\Vx(\bx)+ (\Ex[\bx])^2 \approx \frac{\sum_{i=1}^n(\bx_i-\bar{\bx})^T(\bx_i-\bar{x})}{n-1} + \bar{\bx}^T\bar{\bx} := \text{DE},
\label{lhs}
\end{align}

\noindent where $\bar{\bx}$ is the mean vector. The right-hand side of equation (\ref{rhslhs}) can be estimated using the proposed generative model with a Monte Carlo sample from $p_\theta(\bz)$ and using both $\Vx(\bx\|\bz)$ and $\Ex(\bx\|\bz)$.

\begin{align}
 \Ez[\Vx(\bx\|\bz) + \Ex(\bx\|\bz)^2] &= \int_z (\Vx(\bx \| \bz = z)+\Ex(\bx\|\bz=z)^2)p(z) dz \nonumber \\ 
&\approx \frac{1}{m} \sum_{i=1}^m \left[\Vx(\bx\|\bz=z_i) + \Ex(\bx\|\bz=z_i)^T\Ex(\bx\|\bz=z_i)\right] \\
&:= \text{FME}, \nonumber
\end{align}

where $z_i \sim p(\bz)$, and $\Ex(\bx\|\bz=z_i)$ and  $\Vx(\bx\|\bz=z_i)$ are expressed as functions $f$. Consequently this estimator relies on both components of the fitted LVGM: $p(\bz)$ and $f(\bz)$. This is the forward model estimate (FME). Notice that this estimator does not require we sample from $p_{\theta} (\mathbf{x} \| \mathbf{z} )$ and directly uses the estimated parameters of the emission distribution. It is faster to sample a large amount of $\bz$ than sample a large amount of $\bx$ because traditionally $M << D$. Additionally, this is a simple Monte Carlo sample and it is unbiased\cite{rosenthal19}.

\bigskip

Given a sample of $\Ex(\bx\|\bz=z_i)$ and  $\Vx(\bx\|\bz=z_i)$, it takes $D^2+m$ operations to compute the FME. Based on Shalev-Shwartz and Ben-David's definition of efficiency \cite{shalev2014}, computing the FME would be considered efficient with respect to $D$ and $m$ as the number of operations required is a polynomial function of those parameters, opposed to an exponential function which would be considered inefficient. 

\bigskip

Thus it follows from equation (\ref{rhslhs}) that
\begin{align}
(\Vx(\bx)+ (\Ex[\bx])^2) -  (\Ez[\Vx(\bx\|\bz)] + \Ez[\Ex(\bx\|\bz)^2]) = 0,
\label{eq0}
\end{align}
and since both the DE and the FME are unbiased estimators of the second moment then consequently, the gap between those two estimators reflects how the LVGM captured the second moment of the data set; the bigger the gap is, the poorer the fit is. Thus, we propose to analyse the following moment estimator gap
\begin{align}
\text{DE}-\text{FME} = \text{2MEGA},
\label{megasimple}
\end{align}
which is a matrix of dimension $D \times D$.

\subsection{First Moment}

Similarly, for the first moment we have
\begin{align}
\Ex[\bx] = \Ez[\Ex(\bx\|\bz)],
\label{1stm}
\end{align}
where we estimate the left-hand side with $\bar{\bx}$ (DE) and the right-hand side with $\frac{1}{m} \sum_{i=1}^m \Ex(\bx\|\bz=z_i)$ where $z_i \sim p_\theta(\bz)$ (FME). At this moment, we proposed looking at the gap between moment estimators for both the first and the second moment. However, with machine learning models being mostly optimized towards point estimation, the first moment estimator gap is usually less interesting than its second moment counterpart. 

\subsection{Additional Justifications}

For any generative models, it is always possible to generate a new sample of $G$ points say $S_{LVGM} = \{\tilde{\bx}_1,...,\tilde{\bx}_{G} \}$ and a simple model estimator of the first and second moment would be to simply compute $\frac{1}{G}\sum_{i=1}^G(\tilde{\bx}_i)$ for the first moment and similarly $\frac{1}{G}\sum_{i=1}^G(\tilde{\bx}_i\tilde{\bx}_i^T)$ for the second moment. Let us call these \textit{sample estimators} (SE). These estimators could replace both FMEs defined previously as they also reflect the distribution learned by the LVGM. Conceptually, the SEs are simple and easy to use and metrics such as the MMD relies on additional samples; thus we want to justify why our estimators (FMEs) are \textit{better}.

\bigskip

To begin, FMEs are faster to compute. In order to draw the same sample size for the Monte Carlo estimates, say a sample of size $m$, we need to sample $m$ times for the FMEs. However, for the SEs have to draw twice as many samples ($2m$), since we must sample from $p(\bz)$ $m$ times and then from $p(\bx\|\bz)$ an additional $m$ times. More importantly, because $M << D$ it means that not only FMEs require half as many samples, but these samples are from a much lower dimension distribution which further increases the difference in computational cost between the FMEs and the SEs. 

\bigskip

Another reason why we prefer FMEs over SEs is that FMEs have a smaller variance for both the first and the second moment. Both FMEs and SEs are unbiased estimators of the same value, but the lower variance of FMEs is a huge benefit. A complete proof is located in the appendices. 

\bigskip

\begin{figure}[H]
\centering
\includegraphics[width=11cm]{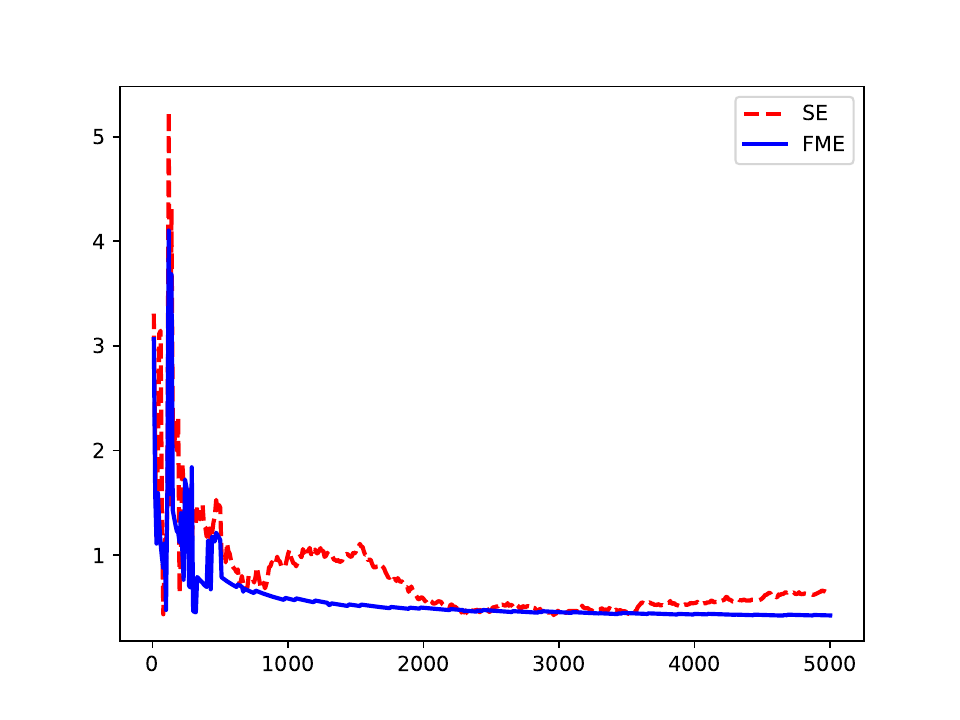}
\caption{The evolution of both gaps plotted against $m$. These gaps are computed using the Frobenius norm as justified in section \ref{mega}.\label{FMEvsSE}}
\end{figure}

Finally, we demonstrated with a simulation that over multiple Monte Carlo samples sizes, the FMEs are closer to the true LVGM moments than SEs. For simple models, such as GMMs, we can compute analytically $ \Ez[\Vx(\bx\|\bz) +$ $\Ex(\bx\|\bz)^2]$, the LVGM second moment. We compared the gap between the true LVGM second moment and the FME to the gap between the true LVGM second moment and the SE and plotted both against $m$ the number of Monte Carlo samples. We can see in figure \ref{FMEvsSE} that the estimator we proposed (FME) is overall closer to the true LVGM second moment than the SE while being faster to compute.

\section{MEGA: A New Metric For Comparing LVGMs} \label{mega}

For latent variable models as defined in section \ref{lvgm} we assess the ability of the model to capture the moment of the data set $S$ by comparing the DE with the FME. Since we are looking at the difference between two different moment estimators of the same value, we named this metric the Moment Estimators Gap (MEGA). We study the difference between our two second moment estimators and we refer to this matrix as 2MEGA. Similarly, we refer to the vector representing the first moment estimator gap as 1MEGA.

\subsection{Selecting A Matrix Norm}

In order to make the MEGA tangible and comparable, we propose to use a matrix norm of the MEGA as our metric. There exist a wide range of possible candidates. Let us introduce a few and justify our finale choice. 

\bigskip

We want to use a norm that looks at the global properties of the matrix and fortunately Rigollet \cite{rigollet15} introduces and studies the behaviour of well-established matrix norms. To begin, we use norms inspired by vector norms. Given the vector $v$, assume $\|v\|_q$ is the following vector norm
\begin{align}
\|v\|_q = (\sum_i \|v_i\|^q)^{(1/q)},
\label{vecnorm}
\end{align}
and given the matrix $M$, its matrix equivalent is $\|M\|_q$
\begin{align}
\|M\|_q = (\sum_{ij} \|M_{ij}\|^q)^{(1/q)}. 
\label{matnorm}
\end{align}

When $q=2$, this is a special case call the Frobenius norm
\begin{align}
\|M\|_2 = \|M\|_F = (\sum_{ij} \|M_{ij}\|^2)^{(1/2)} = \sqrt{\text{Tr}(M^TM)},
\label{Frobe}
\end{align}
where Tr is the trace operator that sums the elements of the diagonal of the input matrix. 

\bigskip

This norm is also a member of the Schatten $q$-norms (for $q=2$) which is a family of matrix norms defined as the vector norm of equation \ref{vecnorm} for the singular values of the matrix. Since we work with second moment estimators, our matrix $M$ is a squared matrix (dimension $m \times m$) and symmetric and thus the singular values of $M$ are equal to its eigenvalues.  We identify the vector of eigenvalues as $\lambda$. Consequently, the Schatten $q$-norm for matrix $M$ is $\|\|M\|\|_q=\|\lambda\|_q$. 

\bigskip

Another member of this family is considered, when $q=\infty$ we define $\|\|M\|\|_\infty = \lambda_{max} = \|\|M\|\|_{op}$ and this is referred to as the operator norm.

\bigskip

In random matrix theory and in matrix estimation, these norms appear frequently and are consequently well known in the statistical research community. For instance, in covariance matrix estimation it is possible, under \textit{mild} assumptions, to bound the operator norm of the difference between the true covariance matrix and simple estimators \cite{rigollet15}. Because of the popularity and the known properties of both the Frobenius norm and the operator norm, they are both legitimate options to measure the MEGA.

\bigskip

The bigger the 2MEGA is, the further away our model second moment is from the data second moment. However, one can perceive the Frobenius norm as the length of the hypotenuse of a multidimensional triangle whose sides are given by the eigenvectors of the matrix while the operator norm is the length of the longest cathetus of the same multidimensional triangle. For computational reasons expressed in the next section, we selected the Frobenius norm and this will serve as our metric in the sections to come. In other words, the metric we propose to evaluate the quality of the fit for the second moment is
\begin{align}
\text{2MEGA-F} &:= \|\text{2MEGA}\|_F.
\label{2MEGAs}
\end{align}

\bigskip

Additionally, if we also consider 1MEGA, we can then again simply use the Frobenius norm (the vector q-norm with $q=2$) on the vector 1MEGA
\begin{align}
\text{1MEGA-F} := \|\text{1MEGA}\|_2.
\label{1MEGAs}
\end{align}

\subsection{Implementation Of The Selected Norm}

Implementing the Frobenius norm is quite straight forward. To compute to Frobenius norm, we need to square the 2MEGA, which involves $m^2$ operations, and then sum its diagonal which results in $m^2+m$ operations. The largest eigenvalue of a matrix can be estimated with the von Mises iteration \cite{mises29}, where each of the $p$ iterations will require $m^2$ operations resulting in $pm^2$ operations in total. Once again, based on Shalev-Shwartz and Ben-David's definition of efficiency \cite{shalev2014}, computing the Frobenius norm of the MEGA would be considered efficient with respect to $m$. Because we can implement an exact computation of the Frobenius norm and because it is computationally faster than the operator norm (as soon as $p \geq 2$) we strictly consider the Frobenius norm for the rest of this article. However, there could be some merit to exploring the operator norm in future work.

\bigskip

We implemented the vector Frobenius norm and the matrix Frobenius norm in Python using the NumPy library \cite{harris20} and using the Pytoch library \cite{Pytorch}. We also implemented a MEGA function that takes as input a sample of $\Vx(\bx\|\bz)$ and $\Ex(\bx\|\bz)$ alongside the data set $S$ and returns 1MEGA-F and 2MEGA-F. These implementations are publicly available on the author's GitHub \cite{beaulac21b} .

\section{Experiments : Comparing Models} \label{expe1}

Now that we established the metric, let us use it to compare distributions $\hat{p}(\bx)$ learned from different models.  

\bigskip

For this demonstration we use simple 2-dimensional observations. We demonstrate that large MEGAs are associated with poor fit when visualizing generated data from LVGMs. We also show that it is concordant with the MMD in most cases. In other words, our metric is concordant with the currently used techniques while providing a new perspective.

\begin{figure}[H]
\centering
\begin{subfigure}{.5\textwidth}
\centering
\includegraphics[width=6cm]{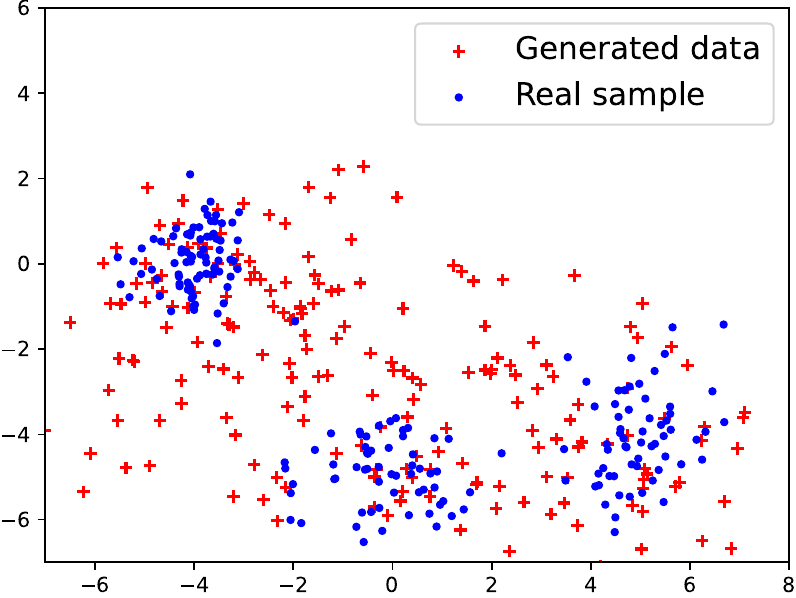}
\caption{Model 1}
  \label{fig1:sub1}
\end{subfigure}%
\begin{subfigure}{.5\textwidth}
\centering
\includegraphics[width=6cm]{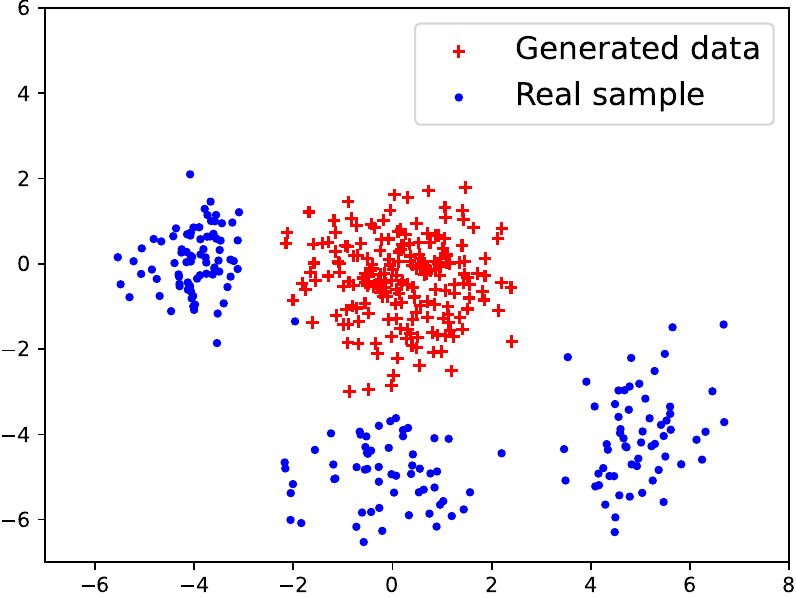}
\caption{Model 2}
  \label{fig1:sub2}
\end{subfigure}
\begin{subfigure}{.5\textwidth}
\centering
\includegraphics[width=6cm]{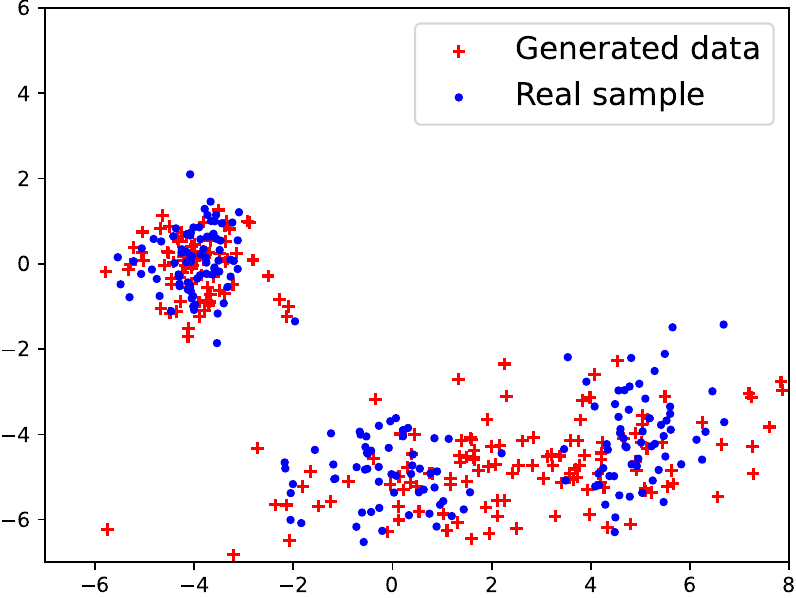}
\caption{Model 3}
  \label{fig1:sub3}
\end{subfigure}%
\begin{subfigure}{.5\textwidth}
\centering
\includegraphics[width=6cm]{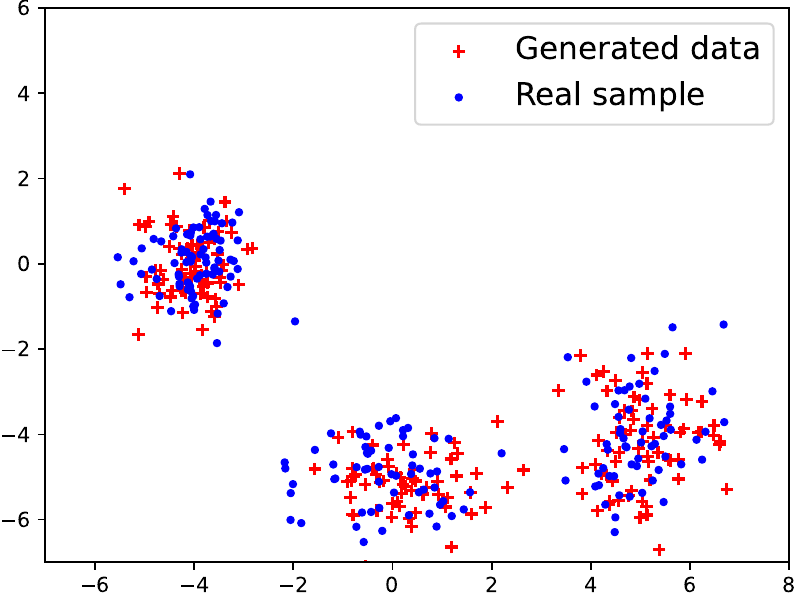}
\caption{Model 4}
  \label{fig1:sub4}
\end{subfigure}
\caption{Generated data from four different LVGM models trained on the three clusters data set.}
\label{BlobRegu}
\end{figure}

\begin{table}[h]
\centering
\begin{tabular}{ |c|c|c|c| }
\hline
 & 1MEGA-F  & 2MEGA-F  & MMD  \\ 
 \hline
Model 1 & 0.2704 & 2.0392 & 0.0812 \\  
Model 2 & 2.4222 & 4.5510 & 1.6717 \\ 
Model 3 & 3.1166 & 3.1712 & 0.0069 \\
Model 4 & 0.2974 & 1.7014 & 0.0012 \\
\hline
\end{tabular}
\caption{MEGA and MMF for the four LVGM models trained on the three clusters data set. \label{BlobRegu_Table}}
\end{table}

Figure \ref{BlobRegu} contains both the training observations (in blue) and generated data points (in red) from four LVGMs. Table \ref{BlobRegu_Table} lists the 1MEGA-F, the 2MEGA-F and the MMD for all four models. We notice first that the 2MEGA-F and the MMD agree on the worst performing model being model 2. A visual inspection of figure \ref{BlobRegu} does support that model 2 fits the data rather poorly. Similarly we notice that 2MEGA-F and MMD agree on the best performing model being model 4 which is also supported by a visual inspection of figure \ref{BlobRegu}. We can also use both MEGA metrics to better understand specific problems with fitted models. For instance, even though a visual inspection of model 1 reveals a poor fit, a small 1MEGA-F tells us the problem is not coming from a poor estimation of the expectation of $\bx$. As a matter of fact, the 2MEGA-F and the MMD are also rather small, indicating the bad fit of model 1 is not due to a poor estimation of the variance either. It is most likely caused by model 1 being unimodal, something that is not measured in the metric we currently propose.

\bigskip

\begin{figure}[H]
\centering
\begin{subfigure}{.5\textwidth}
\centering
\includegraphics[width=6cm]{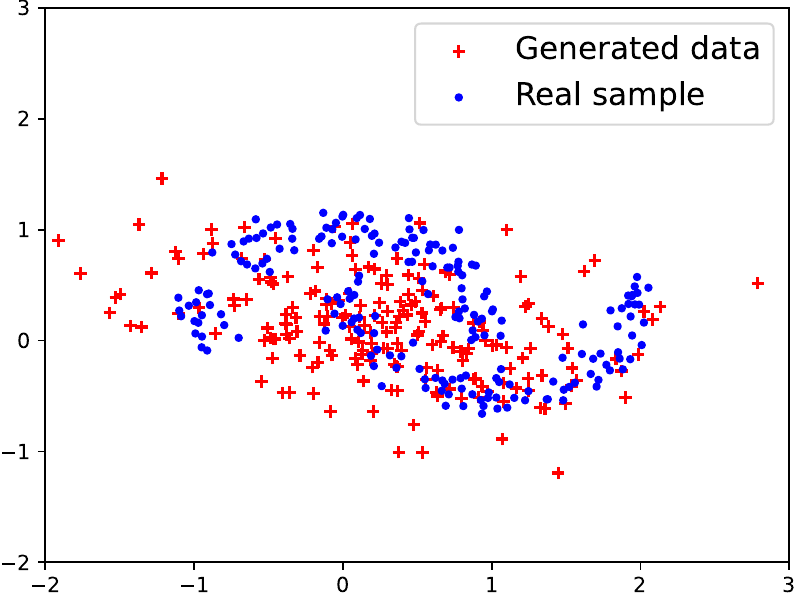}
\caption{Model 1}
  \label{fig2:sub1}
\end{subfigure}%
\begin{subfigure}{.5\textwidth}
\centering
\includegraphics[width=6cm]{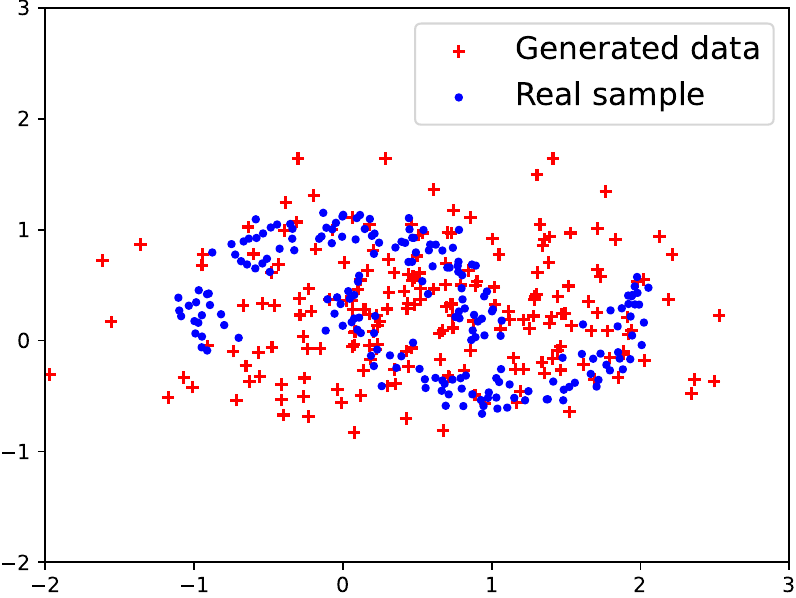}
\caption{Model 2}
  \label{fig2:sub2}
\end{subfigure}
\begin{subfigure}{.5\textwidth}
\centering
\includegraphics[width=6cm]{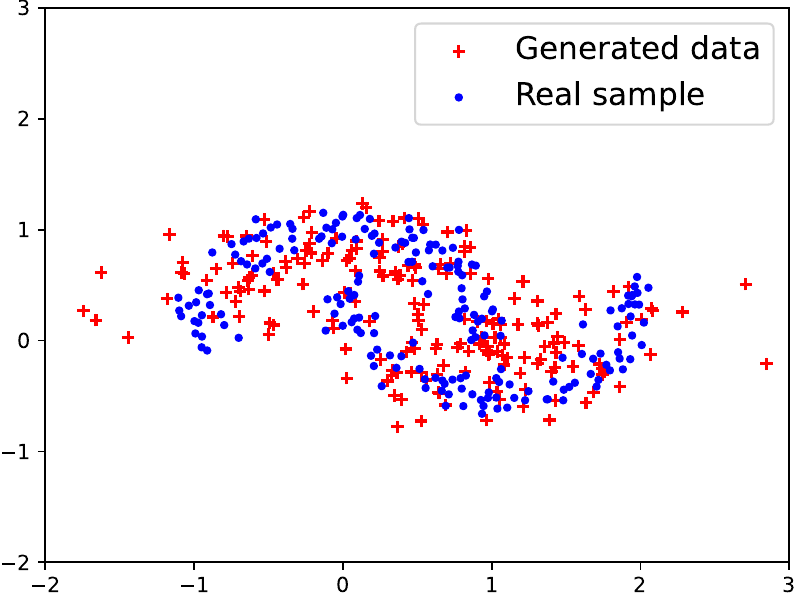}
\caption{Model 3}
  \label{fig2:sub3}
\end{subfigure}%
\begin{subfigure}{.5\textwidth}
\centering
\includegraphics[width=6cm]{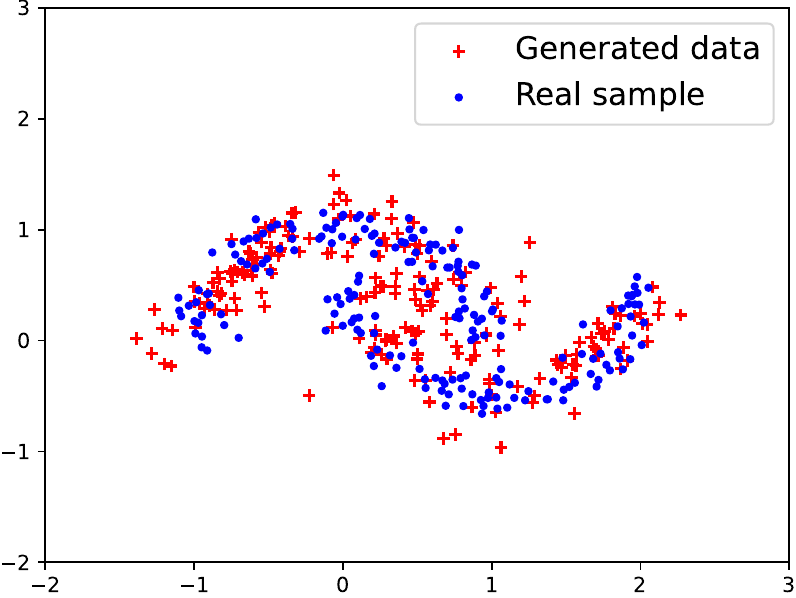}
\caption{Model 4}
  \label{fig2:sub4}
\end{subfigure}
\caption{Generated data from four different LVGMs trained on the half-moon data set. }
\label{MoonRegu}
\end{figure}

Next, we propose another small demonstration with a second data set as illustrated in figure \ref{MoonRegu}.

\begin{table}[h]
\centering
\begin{tabular}{ |c|c|c|c| }
\hline
 & 1MEGA-F  & 2MEGA-F  & MMD  \\ 
 \hline
Model 1 & 0.2565 & 0.5633 & 0.0141 \\  
Model 2 & 0.0023 & 0.5248 & 0.0008 \\ 
Model 3 & 0.7616 & 0.8126 & 7.4341e-05 \\
Model 4 & 0.1072 & 0.4851 & 1.2092e-05 \\
\hline
\end{tabular}
\caption{MEGA and MMD for the four LVGMs trained on the half-moon data set. \label{MoonRegu_Table}}
\end{table}

By visually inspecting figure \ref{MoonRegu}, model 4 seems like a good contender for best fit. In table \ref{MoonRegu_Table}, Model 4 has indeed the lowest 2MEGA-F, which demonstrates again that our proposed metric behaves somewhat intuitively. Additionally, this is concordant with the MMD for which model 4 is also the best model. Visually comparing model 1 and model 2 would be a difficult task since they both seem equally good or equally bad. However, the proposed metric and the MMD are in agreement that model 1 is worse than model 2. Finally, model 3 requires a bit more attention. Even though a visual inspection seems to be positive towards that model it has a rather large first moment estimator gap. Indeed, a closer examination of figure \ref{fig2:sub3} reveals that the centre of mass of the fitted model is at the right end tip of the top half moon rather than in the centre of both half-moons. A careful examination of the metric we propose can help us identify good models, but also reveal aspects of fitted models that require more attention. 

\bigskip

Finally, we provide an additional experiment using real data this time. In this last demonstration, we analyse the well-known MNIST data set \cite{Lecun98} in an unsupervised way. A sequence of GMMs was trained with $K=\{2,4,6,8,10,12,14\}$. We expect intuitively $K=10$ to be a good choice, since the data contains 10 different digits. 

\begin{table}[h]
\centering
\begin{tabular}{ |c|c|c|c| }
\hline
K &  1MEGA-F & 2MEGA-F & MMD \\ 
\hline
2 & 0.24997318 & 2.92840915 & 0.0131 \\  
4 & 0.26690068 & 3.03135908 & 0.0111 \\  
6 & 0.2578028 & 2.925411 & 0.0121 \\  
8 & 0.26219767 & 3.00054297 & 0.0108 \\ 
\hdashline
10 & \textbf{0.23850984} & \textbf{2.82369332} & 0.0089 \\  
\hdashline
12 & 0.26362526 & 3.09359567 & 0.0104\\  
14 & 0.24450328 & 2.92585433 & \textbf{0.0084}\\  
\hline
\end{tabular}
\caption{MEGA and MMD for the sequence of GMM models trained on the MNIST data set. \label{MNIST_Regu}}
\end{table}

In table \ref{MNIST_Regu}, we see that the MEGAs and the MMD agree that the models with 10 and 14 components provide the two best fits of the data. Moreover, our proposed metric favours the model with 10 components which is intuitively a good choice. A natural consequence of using a metric for model comparison is to use it for model selection. In this experiment, we would choose the model with 10 components. In section \ref{regu}, we explore that concept further by discussing the use of the proposed metrics for regularization.

\section{MEGA For Regularization} \label{regu}

As we previously mentioned, based on the principle of Goodhart's Law, we proposed to use the proposed metrics 1MEGA-F and 2MEGA-F to compare complex LVGMs such as IAF-VAEs \cite{kingma16} or NVAEs\cite{vahdat20}. In section \ref{rewo} we discussed some issues with the likelihood as an evaluation metric in some special cases. Additionally, it is unfair to use likelihood as a metric to compare two models when one is trained using a likelihood approach and the other is not.

\bigskip

However, if this metric has some merit when comparing models, can we use it for model selection and can we directly integrate it in the optimization process ? We answer these questions in this section.

\bigskip

Because the likelihood and the moments reflect different aspects of the distribution $p(\bx)$ then incorporating MEGA as part of the optimization process for models trained by maximum likelihood acts as regularization. 

\bigskip

This is easy to see for models with Gaussian emission distribution such as GMMs and VAEs.  When fitting a single Gaussian distribution with maximum likelihood, we set its parameters $\mu$ and $\sigma$ to the data mean and the data standard error and thus a single Gaussian performs very well according to the MEGAs. As we increase the number of components $K$ in a GMM, we increase its likelihood but we also increase its MEGA. This means that we can use the MEGA as a model selection (or regularization metric) to fit GMM in place of alternatives such as AIC and BIC in order to get a model that balances likelihood and moment matching.

\bigskip

One way to do so is to select the model that maximizes:

\begin{align}
\text{ll}(S) - \alpha(\text{1MEGA-F} + \sqrt{\text{2MEGA-F}}),
\label{GMM_regu}
\end{align}
where $\text{ll}(S)$ is the log-likelihood of the data $S$ under the fitted model and $\alpha$ is a hyper-parameter that interpolates between maximum-likelihood and moment estimator. Similar to what is produced when using Lasso \cite{tibshirani96,Simon11,Friedman10} we can draw an entire regularization path (complexity selection path) by evaluating the regularized likelihood of equation \ref{GMM_regu} over different values of $\alpha$.

\bigskip

Similarly, we can use a MEGA term in the objective function when training a VAE. Our proposed metric again serves as a regularizer in order to ensure the VAE model captures those moments by maximizing:

\begin{align}
\textbf{E}_{q}[\ln p(\bx \|\bz)] - KL(q(\bz \| \bx)\| p(\bz)) - \alpha(\text{1MEGA-F} + \sqrt{\text{2MEGA-F}}).
\label{VAE_ruge}
\end{align}

Arguments have been made in the past \cite{Kingma17,higgins17} that the KL divergence term serves as a regularizer and that adding a $\beta$ parameter can provide a way to adjust the strength of the regularization. Adding a MEGA term to the ELBO should provide a different type of regularization. The KL divergence term provides regularization for the distribution $q(\bz \| \bx)$ and the MEGA term provides regularization for $p(\bx)$:

\begin{equation}
\underset{\text{Reconstruction error}}{\textbf{E}_{q}[\ln p(\bx \| \bz)]} - \underset{\text{Regularization for }q(\bz \| \bx)}{\beta KL(q(\bz \| \bx)\|p(\bz))} - \underset{\text{Regularization for }p(\bx)}{\alpha (\text{1MEGA-F} + \sqrt{\text{2MEGA-F}})}.
\label{VAE_ruge2}
\end{equation}

As previously demonstrated \cite{beaulac21}, tunning a VAE to obtain both good reconstruction and generative performances is difficult. For instance, when using the $\beta$-VAE \cite{higgins17,higgins18} we can make the hyper-parameter $\beta$ arbitrary small and obtain close to perfect reconstruction but this leads to poor generative performance because the distribution $q(\bz)$ used for training would be significantly different from $p(\bz)$ used for generation. The additional term we add to the VAE objective function should help with generative perspectives of VAEs.

\bigskip

Thus, to understand the benefits of integrating this additional term to the objective function we have to understand how different it is from the first term. The first term measures the reconstruction error because the likelihood $p(\mathbf{x} \| \mathbf{z})$ is computed on latent variable $\bz$ sampled from $q(\mathbf{z} \| \mathbf{x})$. In contrast, the MEGA penalty term evaluates the first and second moment of $p(\bx)$ independently from $q(\mathbf{z} \| \mathbf{x})$ but rather based on the generative model where $p(\bx) = \int_{z} p(\bx\|\bz)p(\bz) dz$. We believe this additional constraint in the objective function should lead to samples that matches the true distribution of $p(\bx)$ more closely.

\section{Experiments: Regularization} \label{expe2}

\subsection{Regularizer For GMMs}

\begin{figure}[H]
\centering
\includegraphics[width=10cm]{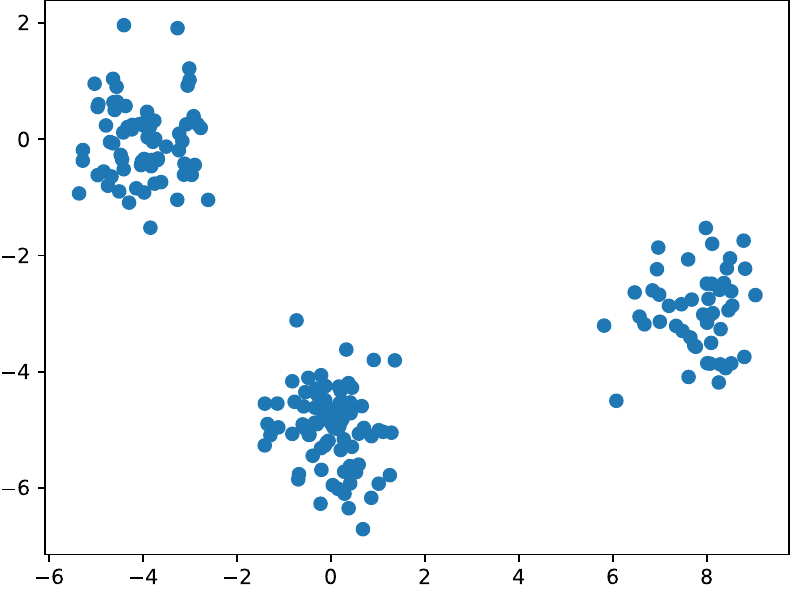}
\caption{Scatter plot of the simulated data set. \label{gmmscat}}
\end{figure}

For GMMs, increasing the number of components increases the likelihood and thus it is necessary to regularize this model; we cannot simply select the ideal number of components based on the likelihood alone. We have generated a simple data set from a 3-component GMM. Figure \ref{gmmscat} is a scatter plot of the simulated data, by looking at the figure we would like the model selection procedure to settle on three components.

The Akaike Information Criteria (AIC) \cite{akaike74,akaike98} is a well-establish penalized likelihood function that can be used to select the number of components in a GMM. The AIC can be expressed as

\begin{align}
\text{AIC} = 2p - 2\text{ll}(S),
\label{AIC}
\end{align}
where $\text{ll}(S)$ is the log-likelihood of the data $S$ under the model, and $p$ is the number of parameters ($p = 2K$ in the GMM case). In equation \ref{AIC} we see a natural tradeoff; $-2\text{ll}(S)$ goes down as we increase the number of components but $2p$ goes up. To select the right GMM model, we can fit multiple GMM with various number of components and we select the model that minimizes the AIC.

\begin{figure}[H]
\centering
\includegraphics[width=10cm]{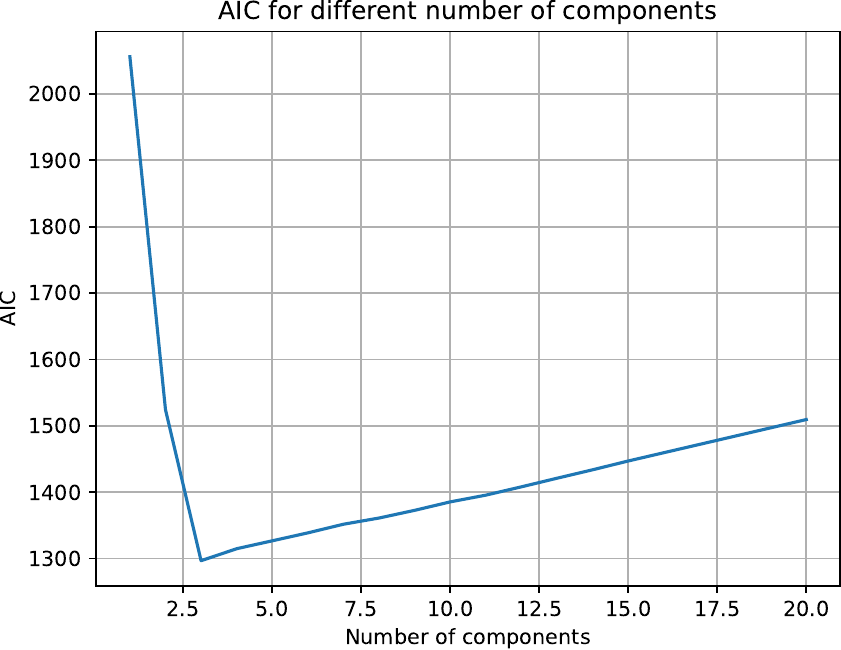}
\caption{AIC plotted against the number of components for trained GMM models (lower is better). \label{aicgmm}}
\end{figure}

Based on figure \ref{aicgmm}, the AIC is minimized at $K=3$ which is concordant the model used to generate the data.

\bigskip

Next, we use our MEGA-penalized likelihood function described in equation \ref{GMM_regu} to select the number of components.

\begin{figure}[H]
\centering
\includegraphics[width=10cm]{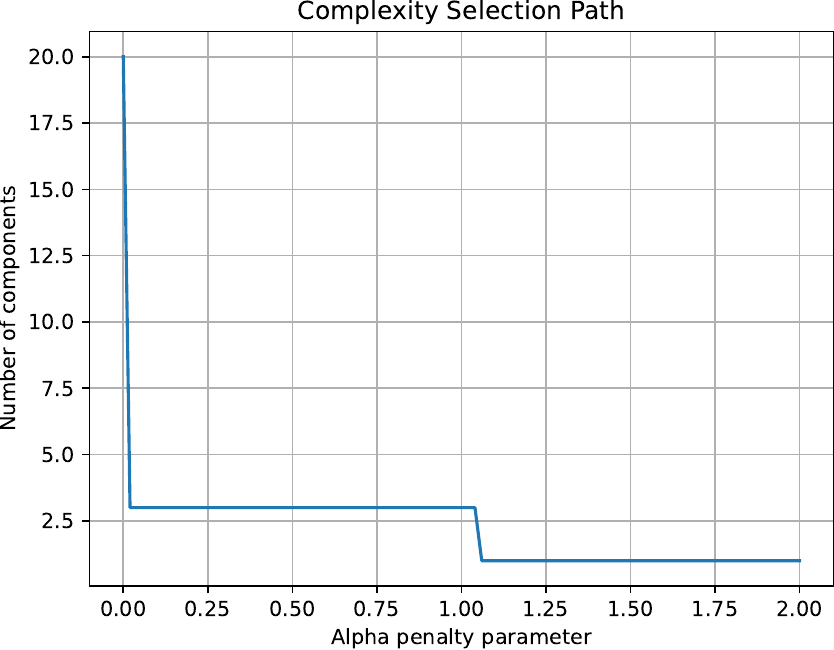}
\caption{Number of components of the selected model plotted against different values of $\alpha$. \label{megagmm}}
\end{figure}

We have computed the MEGA-regularized log-likelihood as expressed in equation \ref{GMM_regu} for different values of $\alpha$. In figure \ref{megagmm}, we plot the number of components of the model that maximizes  the MEGA-regularized log-likelihood for different values of $\alpha$. With small $\alpha$, there is no regularization and the selected model has $K=20$ components which is the highest complexity model fitted. With large values of $\alpha$, the model with the smallest complexity (the model with $K=1$ component) is selected.  In between those two extremes may lie a sequence of nested models and the best one can be selected with the use of a validation set.  In the demonstration, only one model is proposed between the two extreme cases, a model with 3 components which is concordant with both the AIC and the true generative model.  

\bigskip

Compared to the AIC, the flexibility given by the hyper-parameter $\alpha$ is both a benefit and a drawback. Using the hyper-parameter, we can easily adjust how strong we want the penalty to be but on the flip side there are no automated ways to adjust it. 

\subsection{Regularized GMMs For Anomaly Detection}

Because GMMs are popular models for unsupervised tasks, they have been used for anomaly and outlier detection for quite a while now \cite{zimek2012} and are still at the centre of the development of new models such as the deep autoencoding Gaussian mixture model \cite{zong2018}. For outlier detection, we rely on the following procedure. First we fit a GMM to a data set and then point with likelihood lower than a predetermined threshold are considered outliers.

\bigskip

For this demonstration, we use the ionosphere data set \cite{sigillito1989}. This data set consists of 33 variables and 351 observations. This data set has been previously used to test anomaly detection techniques \cite{liu2008,keller2012} and 126 outliers have already been identified.

\bigskip

In this experiment, we select the GMM that maximizes the MEGA-regularized likelihood of equation \ref{GMM_regu} and we then use this model for anomaly detection. For simplicity, we assumed we know there are about 35\% of outliers, this allows us to easily select a threshold that labels the 35\% lowest likelihood data to be outliers. This results in 123 points considered outliers.

\begin{figure}[H]
\centering
\includegraphics[width=10cm]{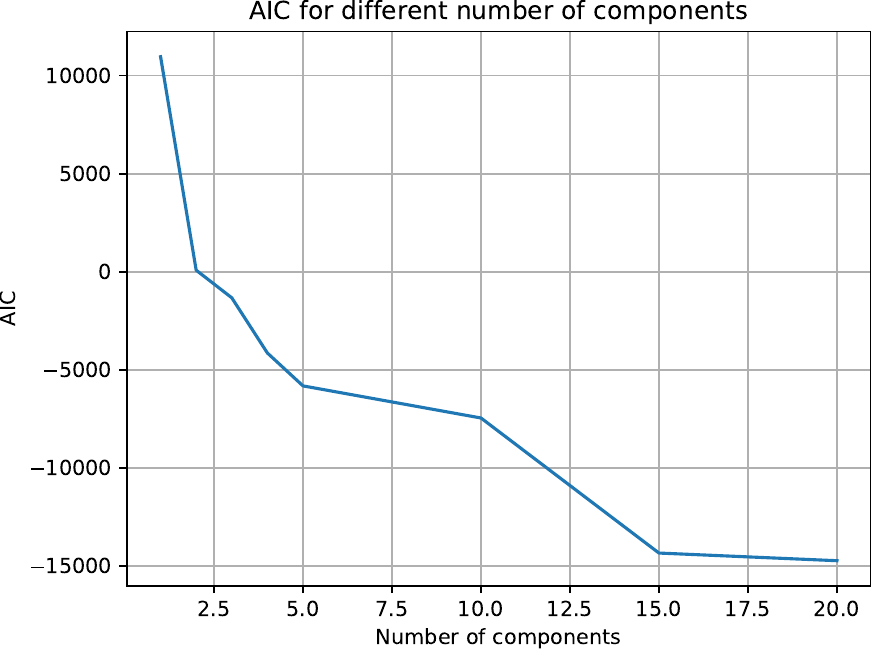}
\caption{AIC plotted against the number of components for trained GMM models (lower is better). \label{aicgmm_anomaly}}
\end{figure}

\begin{figure}[H]
\centering
\includegraphics[width=10cm]{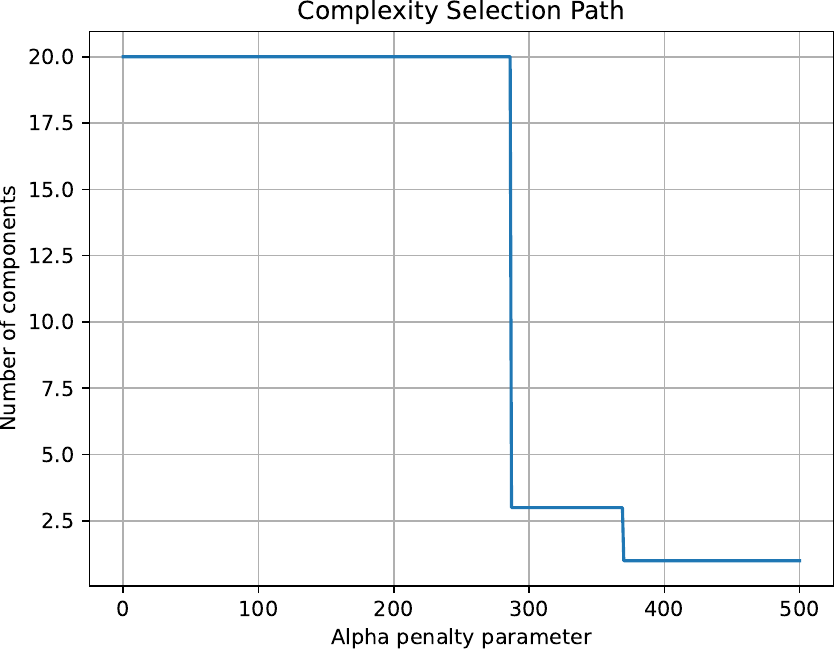}
\caption{Number of components of the selected model plotted against different values of $\alpha$.\label{megagmm_anomaly}}
\end{figure}

Based on figure \ref{aicgmm_anomaly} and figure \ref{megagmm_anomaly}, the AIC suggests a GMM with 20 components and the MEGA-regularized likelihood suggests a model with 3 components.  The model suggested by the AIC correctly identifies 67 outliers, thus missing 59 outliers and miss-labelling 56 observations as outliers. In contrast, the model selected using MEGA correctly identifies 103 outliers while miss-labelling 20 observations as outliers. This also means that this model missed 23 outliers.

\subsection{Regularizer For VAEs}

Finally we experiment with using MEGA as a regularization for VAEs. Now that we used MEGA as part of the objective function we can no longer use it to assess the quality of the samples so we will visually inspect samples and their parameters as well as rely on the Fréchet Inception Distance (FID) \cite{Heusel2017}. 

\bigskip

We run this demonstration on a subset of the MNIST data set \cite{Lecun98} that contains only the digit four. We have experimented with a wide range of parameters $\alpha$ and $\beta$, as defined in equation \ref{VAE_ruge2} and fixed the hyper-parameters to the values leading to the more realistic-looking images. 

\bigskip

\begin{figure}[H]
\begin{subfigure}{.5\textwidth}
  \centering
  \includegraphics[width=.9\linewidth]{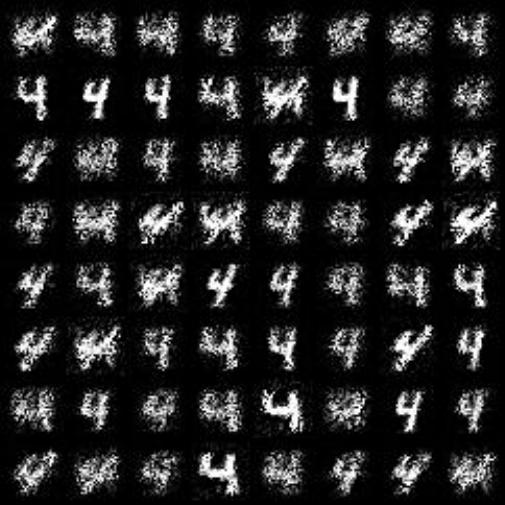}
  \caption{Model train without MEGA}
\end{subfigure}%
\begin{subfigure}{.5\textwidth}
  \centering
  \includegraphics[width=.9\linewidth]{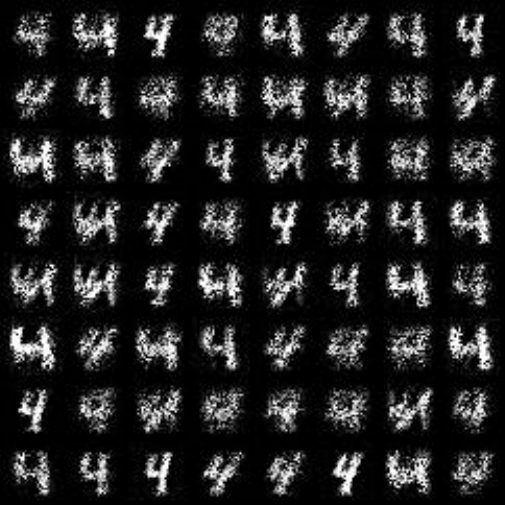}
  \caption{Model train with MEGA}
\end{subfigure}
\label{4lb}
\caption{ A sample of 64 images from $p_\theta(\mathbf{x}\|\mathbf{z}) = N(\mu(\mathbf{z}),\sigma(\mathbf{z}))$ where $\mathbf{z} \sim N(0,1)$. \label{sample_25}}
\end{figure}

\begin{figure}[H]
\begin{subfigure}{.5\textwidth}
  \centering
  \includegraphics[width=.9\linewidth]{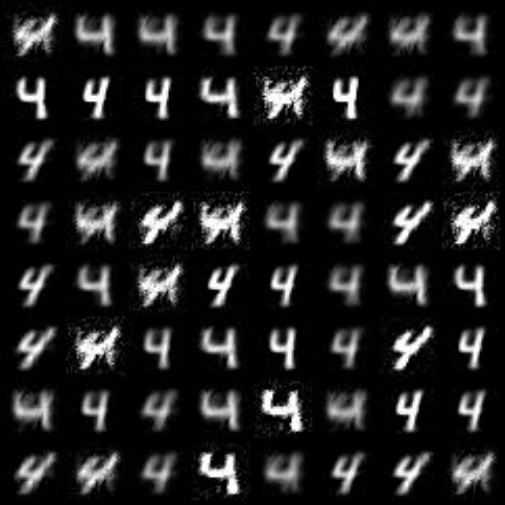}
  \caption{Model train without MEGA}

\end{subfigure}%
\begin{subfigure}{.5\textwidth}
  \centering
  \includegraphics[width=.9\linewidth]{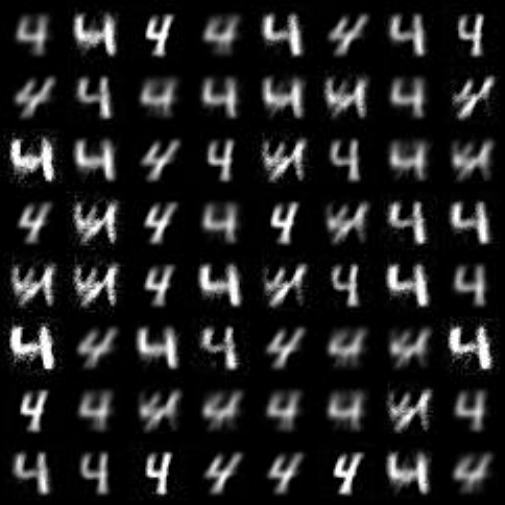}
  \caption{Model train with MEGA}
\end{subfigure}
\caption{ The 64 sampled mean of the images: $\mu(\mathbf{z})$ where $\mathbf{z} \sim N(0,1)$. \label{mu_25}}
\end{figure}

Figures \ref{sample_25} and \ref{mu_25} illustrate our results. The images on in the left column were produced by a VAE trained without MEGA and the right one with MEGA. We have included images of a sample and its mean.

\begin{table}[h]
\centering
\begin{tabular}{ |c|c|c| }
\hline
 & FID for $\bx$ & FID for $\mu(\bz)$ \\ 
 \hline
Without MEGA & 94.9096 &  46.2973 \\  
With MEGA   & 58.3258 &  47.5723\\ 
\hline
\end{tabular}
\caption{FID for VAE and regularized-VAE samples. \label{VAERegu_Table}}
\end{table}

The images themselves are very difficult to analyse which is why we have incorporated table \ref{VAERegu_Table} that contains the FID for both models. Even with the MEGA regularization, the samples are still grainy and this is due to the pixel-independence assumption of the simple VAE which only learns individual pixel variance and not a full-covariance matrix. This is why when data are generated with VAEs, it is common to simply sample $\mu(\bz)$. 

\bigskip

Based on the FID, the images produced with the MEGA-regularized VAE are more realistic. To better understand why that is, we also calculated the FID for the sample of means $\mu(\bz)$. Since those are pretty equivalent for both model, we conclude that the improvement in the image generated with the regularized VAE comes from the captured variance.

\bigskip

In all of the other experiments, the MEGA was computed only a few times, which is extremely fast even on a single CPU. However, for this task, the MEGAs have to be computed for every mini-batch during training. Computing the MEGA over 25 epochs of 20 mini-batch each, for a $D=784$ dimensional observation $\bx$ using a Mont Carlo sample size of $m=5000$ took 1246.0781 seconds on a single CPU (Ryzen 5 5600X). Because the computational time is a polynomial function of these 4 parameters, we believe MEGA can be used as a regularizer on more complex data sets and problems especially if those are trained on clusters of powerful computers.

\section{Discussion}\label{limit}

As a comparative metric, MEGA is fast to compute and easy to interpret. The larger 1MEGA is, the larger the gap is between the first moment of the trained distribution $\hat{p}(\bx)$ and the empirical first moment of the data set $S$. Similarly, the larger 2MEGA is, the larger the gap is for the second moment. This can give us a quick and easy way to compare and evaluate the quality of fitted models.

\bigskip

However, there are still some limitations to this approach. The most obvious is that the formulation we propose only allow us to quickly evaluate the gap for the first two moments. This leads to an incomplete comparison of the learned and empirical distribution which can create some problems in niche cases. A example of this problem is the high performance of the simple Gaussian distribution. Usually, when fitting a Gaussian distribution to a data set we set the parameters $\mu$ and $\sigma$ to be the empirical mean and the empirical standard deviation, thus the Gaussian mixture with a single component shows very good results (low MEGA). 

\bigskip

Fortunately, we have designed this metric for complex latent variable model and there are no reasons to use it when assessing the fit of a single Gaussian. Additionally, other comparative strategies discussed in section \ref{rewo} can still be used in parallel of our proposed one. For instance, the MMD agree with MEGA in most experiments we have done but when they don't agree, they provide different information and can be complementary.

\bigskip

If a trained distribution has low MEGA but bad-looking generated samples, this still provides us with insightful information. It indicates that the problem is in the LVGM distribution's higher moments and our metric was able to provide us with that information very quickly. 

\bigskip

For regularization applications, we were able to successfully used MEGA with two different LVGMs, GMMs and VAEs. Though in both these cases we managed to achieve good results, selecting the appropriate constraint using the hyper-parameters is not an easy task. We recommend plotting the selected number of components against various values of $\alpha$ and to consider all of the models between the most and the least complex model. The use of a validation set can also help select the right value for $\alpha$. 

\bigskip

 We are currently working on a generalization of MEGA. We want to extend our metric not only to higher moments of $\bx$ but to any functions $g(\bx)$. This would not only allow us to compare the skewness of the trained model compared to the data but also more complicated properties of distributions such as multimodality. This generalized MEGA would provide a more complete evaluation of models than the currently proposed metric.

\section{Conclusion} \label{conc}

In this article we introduced a fairly simple and computationally efficient way to check the generative model's distribution of a large class of LVGMs. This metric, MEGA, evaluates the gap between the LVGM distribution's first and second moment and the training (or validation) data set's first and second moment. 

\bigskip

The premise of the proposed metric is theoretically simple and quite intuitive. Both the DE and the FME are unbiased estimators of both the first and the second moment and if a gap exists between them then the LVGM distribution does not match key aspects of the data distribution. 

\bigskip

To support our theoretical arguments, we have demonstrated how to use this metric for two different purposes. First, as an evaluation metric that can replace the more heuristics approaches that rely on eyeballing generated samples. Second, since this metric is currently available for the first two moments, it favours a simple model, such a single Gaussian, and thus can be used as regularization for models such as GMMs and VAEs. 

\bigskip

However, we believe we have only scratched the surface of all of the applications and ways to incorporate these moment-gap-based metrics in model fitting and model selection pipelines. We hope to make further progress in this direction in future work. Another future work direction is to extend these moment gap estimators to sequential LVGMs, such as hidden Markov models and state-space models. Finally, the biggest improvement we could work on is to extend the moment estimators to higher moments; this would make the evaluation metrics much more valuable. 

\bigskip

Nonetheless, we believe this work is a first step in data-driven automated model selection based on moments and we hope it inspires similar contributions.

\section*{Acknowledgement}

I want to thank Yanbo Tang for his valuable help with selecting the matrix norm and for providing sketches for the demonstration that FMEs have smaller variances than SEs. I also want to thank David Duvenaud, Micha\"{e}l Lalancette, Jeffrey S. Rosenthal, Anthony Coache and Renaud Alie for their insightful comments.

\pagebreak

\section*{Appendices}

\subsection*{Proof that FMEs have a small variance than SEs}

\subsubsection*{First moment}

For the SE, we first sample $z_1,...z_m$, a set of $m$ latent variables where $\bz \sim p(\bz)$ then sample $x_1,...,x_m$ with $\bx \sim p(\bx \| \bz_i)$; this implies that $\bx \sim p(\bx \| \bz)p(\bz) = p(\bx)$. Then, the SE for the first moment is $\frac{1}{m} \sum_{i=1}^m \bx_i$.

\begin{align}
\Vx(\text{SE}) = \Vx\left( \frac{\sum_{i=1}^m x_i}{m}\right) =\frac{1}{m^2}\sum_{i=1}^m \Vx(x_i) = \frac{\Vx(\bx)}{m}.
\label{SE1}
\end{align}

When building our FME, we sample $z_1,...z_m$, a set of $m$ latent variables where $\bz \sim p(\bz)$ and then compute $\frac{1}{m} \sum_{i=1}^m \Ex(\bx\|\bz=z_i)$. 

\begin{align}
\Vz(\text{FME}) = \Vz\left( \frac{\sum_{i=1}^m \Ex(\bx\|z_i)}{m}\right) = \frac{1}{m^2}\sum_{i=1}^m \Vz(\Ex(\bx\|z_i)) =  \frac{\Vz[\Ex(\bx\|z_i)]}{m}.
\label{FME1}
\end{align}

Then using the Law of Total Variance we have that:

\begin{align}
\Vx(\bx) &= \Ez[\Vx(\bx\|\bz)] + \Vz[\Ex(\bx\|\bz)] \notag \\
&\geq \Vz[\Ex(\bx\|\bz)] \notag \\
\Rightarrow \frac{\Vx(\bx)}{m} &\geq \frac{\Vz[\Ex(\bx\|z_i)]}{m} \\
\Rightarrow \V(\text{SE}) &\geq \V(\text{FME}) \notag,
\end{align}

\noindent and thus our FME has lower variance than the commonly used alternative.

\subsubsection*{Second moment}

For the SE, we first sample $z_1,...z_m$, a set of $m$ latent variables where $\bz \sim p(\bz)$ then sample $x_1,...,x_m$ with $\bx \sim p(\bx\|\bz_i)$; this implies that $\bx \sim p(\bx\|\bz)p(\bz) = p(\bx)$. Then, the SE for the second moment is $\frac{1}{m} \sum_{i=1}^m \bx_i^2$.

\begin{align}
\Vx(\text{SE}) = \Vx\left( \frac{\sum_{i=1}^m x_i^2}{m}\right) = \frac{\Vx(\bx^2)}{m}.
\label{SE2}
\end{align}

When building our FME, we sample $z_1,...z_m$, a set of $m$ latent variables where $\bz \sim p(\bz)$ and then compute $\frac{1}{m} \sum_{i=1}^m \left(\Vx(\bx\|\bz=z_i) + [\Ex(\bx\|\bz=z_i)]^2\right)$.

\begin{align}
\Vz(\text{FME}) &= \Vz\left( \frac{\sum_{i=1}^m \left(\Vx(\bx\|\bz=z_i) + [\Ex(\bx\|\bz=z_i)]^2\right)}{m}\right) \notag \\ 
&= \frac{\Vz\left(\Vx(\bx\|\bz=z_i) + [\Ex(\bx\|\bz=z_i)]^2\right)}{m}.
\label{FME2}
\end{align}

\noindent Now let's take a closer look at the numerator of equation (\ref{FME2}).

\begin{align}
\Vz\left(\Vx(\bx\|\bz=z_i) + [\Ex(\bx\|\bz=z_i)]^2\right) & = \Vz\left(\Ex[\bx^2\|\bz] - \Ex[\bx\|\bz]^2 + \Ex[\bx\|\bz]^2\right) \notag \\
& = \Vz\left(\Ex[\bx^2\|\bz] - \Ex[\bx\|\bz]^2 + \Ex[\bx\|\bz]^2\right)  \notag \\
& = \Vz\left(\Ex[\bx^2\|\bz]\right) \notag \\
& = \Ez\left(\Ex[\bx^2\|\bz]^2\right)-\Ez\left(\Ex[\bx^2\|\bz]\right)^2 \\
& = \Ez\left(\Ex[\bx^2\|\bz]^2\right)-\Ex\left(\bx^2\right)^2 \notag \\
\Rightarrow \V(\text{FEM}) &= \frac{\Ez\left(\Ex[\bx^2\|\bz]^2\right)-\Ex\left(\bx^2\right)^2}{m} \notag
\label{FME2num}
\end{align}

Finally, let us apply the Law of Total Variance to $\bx^2$:

\begin{align}
\Vx(\bx^2) &= \Ez(\Vx[\bx^2\|\bz])+\Vz(\Ex[\bx^2\|\bz])\notag \\
& = \Ez(\Vx[\bx^2\|\bz])+\Ez\left(\Ex[\bx^2\|\bz]^2\right) - \Ez(\Ex[\bx^2\|\bz])^2 \notag \\
& = \Ez(\Vx[\bx^2\|\bz])+\Ez\left(\Ex[\bx^2\|\bz]^2\right) - \Ex[\bx^2]^2\\
\Rightarrow \Vx(\bx^2) &\geq \Ez\left(\Ex[\bx^2\|\bz]^2\right) - \Ex[\bx^2]^2 \notag \\
\Rightarrow \frac{\Vx(\bx^2)}{m} &\geq \frac{\Ez\left(\Ex[\bx^2\|\bz]^2\right) - \Ex[\bx^2]^2}{m} \notag \\
\Rightarrow \V(\text{SE}) &\geq \V(\text{FME}) \notag
\label{FME2num}
\end{align}

\pagebreak

\bibliography{mybibfile}


\end{document}